\documentclass{esannV2}
\usepackage[dvips]{graphicx}
\usepackage[utf8]{inputenc}
\usepackage{amssymb,amsmath,array}

\usepackage[normalem]{ulem}
\usepackage{bbm}
\newcommand{\ind}{\mathbbm{1}}
\usepackage{comment}
\newcommand{\myboldparagraph}[1]{\smallskip\noindent\textbf{#1}}
\newcommand{\myparagraph}[1]{\smallskip\noindent\emph{#1}}

\newcommand{\specificthanks}[1]{\@fnsymbol{#1}}
\usepackage{url}
\usepackage{booktabs}
\usepackage{xcolor}
\usepackage{subcaption}

\usepackage{amsmath}

\begin{document}

\title{On the Limitations of Model Stealing with Uncertainty Quantification Models}

\author{David Pape$^1$, Sina Däubener$^{2}$\thanks{Funded by the Deutsche Forschungsgemeinschaft (DFG, German Research Foundation) under Germany's Excellence Strategy  - EXC 2092 CASA – 390781972.} \ , Thorsten Eisenhofer$^{2*}$,\\ Antonio Emanuele Cinà$^1$, Lea Schönherr$^1$
\vspace{.3cm}\\
1- CISPA Helmholtz Center for Information Security, 
2- Ruhr University Bochum\\
}

\maketitle

\begin{abstract}
Model stealing aims at inferring a victim model's functionality at a fraction of the original training cost.
While the goal is clear, in practice the model's architecture, weight dimension, and original training data can not be determined exactly, leading to mutual uncertainty during stealing.
In this work, we explicitly tackle this uncertainty by generating multiple possible networks and combining their predictions to improve the quality of the stolen model.
For this, we compare five popular uncertainty quantification models in a model stealing task. %

Surprisingly, our results indicate that the considered models only lead to marginal improvements in terms of label agreement (i.e., fidelity) to the stolen model.
To find the cause of this, we inspect the diversity of the model's prediction by looking at the prediction variance as a function of training iterations. We realize that during  training, the models tend to have similar predictions, 
indicating that the network diversity we wanted to leverage  using uncertainty quantification models is not (high) enough for improvements on the model stealing task.

\end{abstract}

\section{Introduction}

\textit{Machine Learning as a Service} (MLaaS) enables an easy and cost-effective way to develop machine learning services. However, it also increases the risk of model stealing for attackers who can exploit barrier-free invocations, such as APIs~\cite{Cina2022Survey}. %
For this, model stealing aims at inferring the model functionalities from a black-box model at a fraction of the original training costs~\cite{jagielski-20-high} while having access only to the outputs of the black-box model. 
Therefore, the attacker trains a \emph{surrogate model} on a queried dataset. %
Previous studies assumed either full or partial knowledge about the network architecture~\cite{orekondy-19-knockoffnets, kariyappa-21-maze}, which is often already indicative for the network functionalities, or attempted to reverse engineer it through probing~\cite{joon-19-reverse}. However, in practice, the architecture cannot be precisely determined, leading to mutual model~uncertainty.

In this work, we hypothesize that the attacker can explicitly tackle this uncertainty by simultaneously considering multiple networks from the model space, which is naturally done in Bayesian model averaging. %
Therefore, we evaluate five different uncertainty quantification models, which sample from a (learned) parameter distribution during inference, as the surrogate model to reinterpret each sample from the parameter distribution as one possible target network:
\emph{Bayesian Neural Networks}~\cite{https://doi.org/10.48550/arxiv.1505.05424}, \emph{Monte Carlo Dropout}~\cite{https://doi.org/10.48550/arxiv.1506.02142}, \emph{Concrete Dropout}~\cite{NIPS2017_84ddfb34} and the straight-forward but more costly approach of deep and heterogeneous ensembles~\cite{https://doi.org/10.48550/arxiv.1612.01474, lukovnikov-etal-2021-detecting-compositionally}.
To test our hypotheses, we compare the model-stealing performance of the uncertainty quantification models 
with a single deterministic model.
In all approaches, we consider different-sized target models trained for image classification tasks on CIFAR10~\cite{Krizhevsky2009LearningML} and SVHN~\cite{SVHN}.

Our experiments show that uncertainty quantification models only lead to insignificant improvements over a single-model baseline,
implying that the Bayesian model average does not lead to improvements in mimicking the functionalities of the target model.
To gain a deeper understanding of this, we analyzed the variance of the model predictions as a function of training iterations. This shows that during training, the models converge to similar predictions, indicating a limited function variability on the test data. %

In summary, we make the following key contributions:
\begin{itemize}
    \item We present the first evaluation of uncertainty quantification models used in the context of model stealing and evaluate them in terms of fidelity.
    \item We further discuss their limitations %
    by analyzing fidelity in relation to the model's output variance.
\end{itemize}

\section{Background}

All uncertainty quantification methods
in this paper derive their final network predictions based on the following approximation %
\begin{equation}\label{eq:mc_approx}
    f(y|x) \approx \frac{1}{M}\sum_{i=1}^M f(y|x, \theta_i) \text{ with } \theta_i \sim q(\theta) \enspace,
\end{equation}
where $x$ and $y$ are the input and corresponding label, and $\theta_i$ are parameters of the model drawn $M$ times from an underlying distribution.
In the following, we briefly explain the differences of $q(\theta)$ for each network type

\textbf{Bayesian Neural Networks.}
Bayesian neural networks (BNNs) are often referred to as a principal way to quantify uncertainty. One specific characteristic of those networks is the derivation of a posterior distribution $q(\theta)$ over model parameters. In this setting, eq.~\eqref{eq:mc_approx} could be interpreted as the Monte Carlo approximation of the posterior predictive distribution.

\textbf{Monte Carlo Dropout.}
Monte Carlo (MC) Dropout is an (approximate) Bayesian method where neurons are randomly dropped with a fixed dropout probability during training as well as during inference. These pattern of deactivated neurons are named dropout masks. In this setting $q(\theta)$ could be interpreted as the distribution over these dropout masks. %

\textbf{Concrete Dropout.}
Contrary to MC dropout, the dropout probability in concrete dropout~(CD) is learned through a continuous relaxation of the discrete dropout mask. The interpretation of $q(\theta)$ stays nevertheless identical. 

\textbf{Deep Ensembles.}
In Deep Ensembles (DEs) multiple networks with the same network architecture but different initial weight values are trained. When trained with weight decay, these can be seen as samples from a posterior $q(\theta)$. %

\textbf{Heterogeneous Ensembles.}
Going one step further, Heterogeneous Ensembles (HEs) combine different network architectures with different properties to an ensemble enabling a broader exploration of the function space. %

\section{Stealing with Uncertainty Quantification~Models}

\myboldparagraph{Adversary Goal.}
The attacker's goal is to create a surrogate model $\hat{f} $ that maximizes the prediction agreement, referred to as  \textit{fidelity}~\cite{https://doi.org/10.48550/arxiv.1909.01838}, given by

\begin{equation}
    \frac{1}{|D_{test}|} \sum_{x_i \in D_{test}} \ind \{ \hat{f}(x_i) = f(x_i)\} \enspace ,
\end{equation}
with a target model $f$ for a test set $D_{test}$, where $\ind\{ \cdot \}$ is an indicator function. %

We assume that the adversary has knowledge of the semantics of the black-box oracle; that is, they know the target model's input representation and the corresponding task. %
Furthermore, we also assume the attacker has access to public task-relevant pretrained models or datasets.
The adversary has no knowledge of the inner workings of the target model. This includes the architecture, hyperparameters, training procedure, and training dataset. 
Given an image $x \in \mathbb{X}$, the adversary receives a target label $y \in \{0, \dots, k\}$, where k is the number of classes. Furthermore, we assume that the attacker can send unlimited queries to the target and retrieve the corresponding labels~\cite{jagielski-20-high}. %

\myboldparagraph{Experimental Setup.}
All experiments are conducted on the CIFAR10 and SVHN datasets while using four NVIDIA GeForce RTX 2080 Ti.%
We use the first half of the respective test set to evaluate the target models. Fidelity calculations of the surrogate models are conducted on the second half.  %

\myparagraph{Target models.} We consider three different target models of varying sizes. A small and a medium-sized model were trained from scratch. As a large model, we fine-tuned a pretrained ResNet152-V2. 
The training is conducted on half of the training datasets, respectively. We use the categorical cross-entropy loss and the Adam optimizer with an initial learning rate of $1e-5$. The number of parameters and accuracies are reported in Table~\ref{tab:black-boxes}. %

\begin{table}[tb]
    \centering
    \caption{Number of parameters for the target models and their accuracies in \% for CIFAR10 and SVHN.}
    \resizebox{0.7\columnwidth}{!}{
    \begin{tabular}{c|c|c|c}
    \toprule
        Name& Parameters & CIFAR10 Accuracy & SVHN Accuracy  \\ \midrule
        Small& ~~~196,352 & 83.6 & 88.5 \\
        Medium& ~2,040,352 & 88.2 & 92.0\\
        Large& 63,582,218 & 93.7 & 95.6\\
        \bottomrule
    \end{tabular}}
    \label{tab:black-boxes}
\end{table}

\myparagraph{Surrogate models.} For training all surrogate models, the adversary generates a surrogate dataset by querying the target model with the second half of each training dataset.
We used pretrained architectures and finetuned them for 30 epochs for all models except for the BNN where we used 50 epochs because of slower training convergence. We use the following surrogate models:

\texttt{Baseline.} A ResNet152V2 (Res) and an InceptionV3 (Inc) architecture with an added feed-forward classification head. %

\texttt{MC Dropout (MCD).} As an extension of the baseline model, where we added two dropout layers each with a dropout rate of 50\%  before the last two layers in the feed-forward head. %

\texttt{Concrete Dropout (CD).} Modifies the baseline model by replacing all layers of the feed-forward head with concrete dropout layers of the same width. %

\texttt{Bayesian Neural Network (BNN).} The baseline models architecture is altered by replacing  all layers of the feed-forward head by probabilistic reparameterization layers and trained via BayesByBackprob~\cite{https://doi.org/10.48550/arxiv.1312.6114}.

\texttt{Deep Ensemble (DE).} Consists out of six baseline models each with a randomly initialized classification head.

\texttt{Heterogeneous Ensemble (HE).} Combines six different pre-trained model architectures: ResNet50, ResNet152V2, VGG16, VGG19, InceptionV3, and \linebreak DenseNet169. %

For inference, 50 forward passes for the dropout and Bayesian models are~used.

\begin{table}[tb]
    \centering
    \caption{Fidelity of the surrogate models in \% for different network architectures and sizes. Bold numbers highlight the maximum fidelity.}
    \vspace{-0.5em}
    \subfloat[CIFAR10]{
        \resizebox{.9\columnwidth}{!}{%
        \begin{tabular}{c|cc|cccccccccc}
        \toprule
            Target  & \multicolumn{2}{c|}{Baseline}& \multicolumn{2}{c}{MCD} & \multicolumn{2}{c}{CD} & \multicolumn{2}{c}{BNN} & \multicolumn{2}{c}{DE} & HE \\ 
             & Res & Inc & Res & Inc & Res & Inc & Res & Inc & Res & Inc & - \\
            \cmidrule(lr){1-12}  %
            Small & 85.18&	85.14	&85.10	&84.82	&85.48	&84.80	&84.88	&83.24	&85.88	&85.50	&\textbf{86.96}\\ 
            Medium & 88.91	&89.38	&89.86&	89.62	&89.72	&89.52	&88.48	&86.00	&90.04	&90.06	&\textbf{90.49} \\
            Large & 93.72	&92.44&	93.24	&92.18&	93.20	&92.52	&90.72	&87.84	&93.94	&93.04	&\textbf{94.09} \\
            \bottomrule
        \end{tabular}%
        }
        }\\
    \subfloat[SVHN]{
        \resizebox{.9\columnwidth}{!}{%
        \begin{tabular}{c|cc|cccccccccc}
        \toprule
        Target  & \multicolumn{2}{c|}{Baseline}& \multicolumn{2}{c}{MCD} & \multicolumn{2}{c}{CD} & \multicolumn{2}{c}{BNN} & \multicolumn{2}{c}{DE} & HE \\ 
         & Res & Inc & Res & Inc & Res & Inc & Res & Inc & Res & Inc & - \\
        \cmidrule(lr){1-12}  %
        Small & 91.06	&90.89&	90.65&	90.17&	91.33&	90.89&	91.15&	90.88&	92.45&	91.78&	\textbf{92.55}\\ 
        Medium & 92.93	&92.59&	92.61&	92.40&	93.24&	92.29&	92.87&	92.11	&93.68&	93.52&	\textbf{94.08} \\
        Large & 95.74&	94.38&	95.07&	94.42&	95.85&	94.48&	95.06&	94.16&	\textbf{96.52}&	95.65&	95.84 \\
        \bottomrule 
        \end{tabular}%
        }
        }
    \label{tab:results50fp}
\end{table}

\myboldparagraph{Fidelity results.}
From the fidelity of the different surrogate methods and architectures in Table~\ref{tab:results50fp}, 
it can be seen that a ResNet152V2 surrogate architecture leads to improved fidelity for the large target model. This indicates that a higher degree of similarity between the target and surrogate architecture positively impacts the effectiveness of model stealing. Conversely, other architectures only marginally influence fidelity.
We further note %
that the BNN does not increase the fidelity for any target model in comparison to the baseline model. Similarly, MC dropout produces only minor improvements for stealing the medium target model using CIFAR10. %
Improvements over the baseline for model stealing with CD can be seen for several combinations of target size and model architectures for both datasets. However, an ensemble of models consistently improves fidelity, while specifically HEs consistently reach the highest fidelity for~CIFAR10.

\myboldparagraph{Vanishing prediction variance.}
\begin{figure}[t]
    \centering
    \subfloat[CIFAR10 and ResNet152V2]{
        \includegraphics[width=0.49\textwidth]{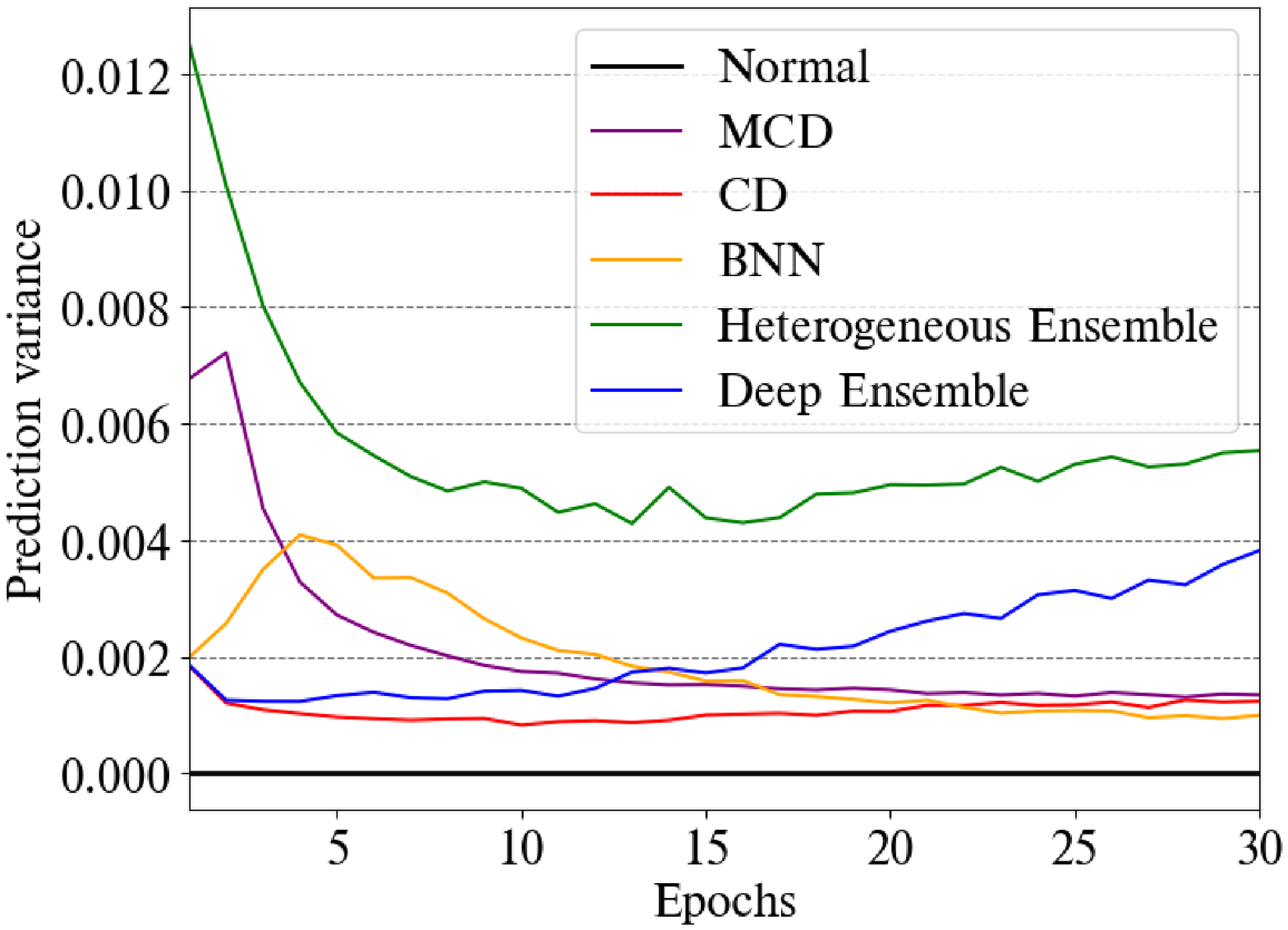}
    }
    \subfloat[SVHN and InceptionV3]{
        \includegraphics[width=0.49\textwidth]{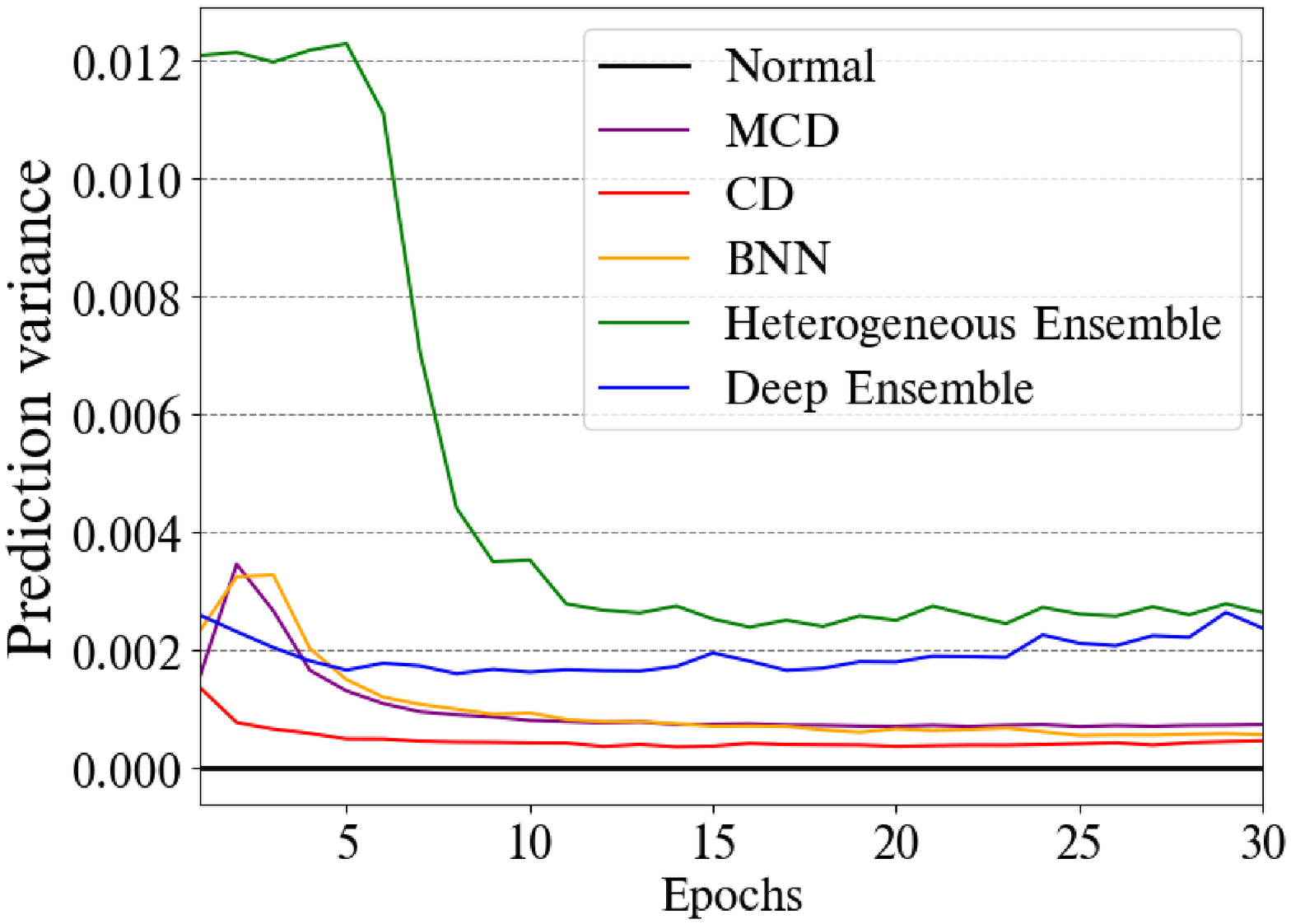}
    }
    \vspace{-0.5em}
  \caption{Variance of the predictions generated by the subnetworks for each method plotted over the course of training. We show the variances for two combinations of dataset and surrogate architecture.}
  \label{fig:variances}
\end{figure}
Our initial experiments raise the question of why uncertainty quantification models only slightly improve the performance of model stealing attacks. We hypothesize, that the induced network variability through sampling does not lead to very diverse networks, such that they are not able to eliminate potential failure cases. To test this, we show in Figure~\ref{fig:variances} the prediction variance calculated on the test set generated by the small target model over the course of the training epochs for all architectures. A lower variance in the different output predictions stems from the same/similar predictions of all subnetworks, which indicates a weak model exploration of subnetworks on these datapoints. 
We observe that the prediction variance for MC dropout and the BNN first increases, probably due to an initial warm-up phase with small weights. In the further course, the variance decreases to less than 0.002. This level is also not surpassed by CD. Furthermore, we observe that the prediction variance of our DE increases during training. This could be traced back to findings from Fort et al.~\cite{https://doi.org/10.48550/arxiv.1912.02757}, where the authors show that ``deep ensemble tend to explore multiple modes in function space'', whereas BNNs often focus on a single mode, leading to less variability. Compared to the others, only the HE has a notably different prediction variance. Note, that the HE uses different model architectures and different pre-training, while only the feed-forward heads in the Bayesian, dropout, and deep ensembles models induce variability. 
Consequently, heterogeneous models in an ensemble preserve higher prediction variance, indicating higher function space diversity which could be the cause for their  improved fidelity.

This diminishing variance over the course of training leads
us to test the practical implications of the amount of forward passes during the prediction of the Bayesian and dropout models. Hence, we reduced the
amount of forward passes to six, the same number of sub-models we used in the
ensembles. We observe that indeed, the amount of forward passes has little to no impact on the final fidelity of the surrogate models. \\
Furthermore, we assess the training cost of uncertainty quantification models and compare it with respect to the training cost of a standard model. Our results show that all methods result in much higher training and prediction times due to the significant changes to the training and inference procedure. This means using uncertainty quantification models results in a significant time overhead compared
to standard models.

\section{Conclusion}
The availability of MLaaS and the high costs of training ML models demonstrate model stealing as a considerable security threat.
In practice, however, the architecture of a black-box model and the characteristics of the weights cannot be determined precisely, introducing an inherent uncertainty for a successful extraction.
In this work, we explicitly tackle this uncertainty when staging a model-stealing attack by using models for uncertainty quantification, which allow the attacker to simultaneously probe %
multiple network configurations.

Our findings demonstrate that, in general, this approach only leads to \linebreak marginal improvements.
Furthermore, we have shown that it is difficult to maintain a high model variability for increasing training epochs.
Lastly, we observe that combining different architectures into an ensemble can slightly improve upon the baseline, even if the latter uses the same architecture as the victim~model.

\section*{Acknowledgment}
This work has been partially supported by Spoke 10 ``Logistics and Freight'' within the Italian PNRR National Centre for Sustainable Mobility (MOST), CUP I53C22000720001.

\begin{footnotesize}

\bibliographystyle{unsrt}
\bibliography{main}

\end{footnotesize}

\end{document}